\documentclass[10pt,twocolumn,letterpaper]{article}

\usepackage{iccv}
\usepackage{times}
\usepackage{epsfig}
\usepackage{graphicx}
\usepackage{amsmath}
\usepackage{amssymb}

\usepackage[utf8]{inputenc}
\usepackage{subcaption}  
\usepackage{blindtext}
\usepackage{pgfplots}
\pgfplotsset{compat=1.14}
\usepackage{tikz}
\usepackage{tikzscale}  
\usepackage{color, colortbl}  
\usepackage{pifont}  
\usepackage{diagbox} 
\usepackage{makecell} 
\usepackage{float}
\usepackage{authblk}
\usepackage{siunitx}
\usepackage{bm}
\usepackage{placeins}

\usepackage[pagebackref=true,breaklinks=true,letterpaper=true,colorlinks,bookmarks=false]{hyperref}

\iccvfinalcopy 

\def\iccvPaperID{5626} 

\ificcvfinal\fi

\newcommand\fe[0]{{f.~e.~}}
\newcommand\figref[1]{Fig.~{\ref{#1}}}
\newcommand\tabref[1]{Table~{\ref{#1}}}
\newcommand\etalcite[1]{{\etal~\cite{#1}}}
\newcommand{\norm}[1]{\left\lVert #1 \right\rVert_2}
\DeclareMathOperator*{\argmin}{arg\,min}

\newcommand\Y{\textcolor{green}{\ding{51}}}
\newcommand\N{\textcolor{red}{\ding{55}}}

\newcommand\ourdb{\text{FreiHAND}}

\newcommand{\cmt}[1]{}  
\begin{document}

\title{\vspace*{-1.5em}\ourdb: A Dataset for Markerless Capture of Hand Pose and Shape from Single RGB Images\vspace*{-1.5em}}

\author[1]{Christian Zimmermann}
\author[2]{Duygu Ceylan}
\author[2]{Jimei Yang}
\author[2]{Bryan Russell}
\author[1]{\authorcr Max Argus}
\author[1]{Thomas Brox}
\affil[1]{University of Freiburg}
\affil[2]{Adobe Research}
\affil[ ]{\textit{ \small \mbox{\small Project page: \url{https://lmb.informatik.uni-freiburg.de/projects/freihand/}} }}

\maketitle

\begin{abstract}
Estimating 3D hand pose from single RGB images is a highly ambiguous problem that relies on an unbiased training dataset. In this paper, we analyze cross-dataset generalization when training on existing datasets. We find that approaches perform well on the datasets they are trained on, but do not generalize to other datasets or in-the-wild scenarios. As a consequence, we introduce the first large-scale, multi-view hand dataset that is accompanied by both 3D hand pose and shape annotations. For annotating this real-world dataset, we propose an iterative, semi-automated `human-in-the-loop' approach, which includes hand fitting optimization to infer both the 3D pose and shape for each sample. We show that methods trained on our dataset consistently perform well when tested on other datasets. Moreover, the dataset allows us to train a network that predicts the full articulated hand shape from a single RGB image. The evaluation set can serve as a benchmark for articulated hand shape estimation. 
\end{abstract}

\section{Introduction}
3D hand pose and shape estimation from a single RGB image has a variety of applications in gesture recognition, robotics, and AR. Various deep learning methods have approached this problem, but the quality of their results depends on the availability of training data. Such data is created either by rendering synthetic datasets~\cite{boukhayma20193d, ge20193d, mueller2018ganerated, OccludedHands_ICCV2017, zb2017hand} or by capturing real datasets under controlled settings typically with little variation~\cite{gomez2019large, simon2017hand, tzionas2016capturing}. Both approaches have limitations, discussed in our related work section.

\begin{figure}[ht]
\centering
    \includegraphics[width=1.0\columnwidth]{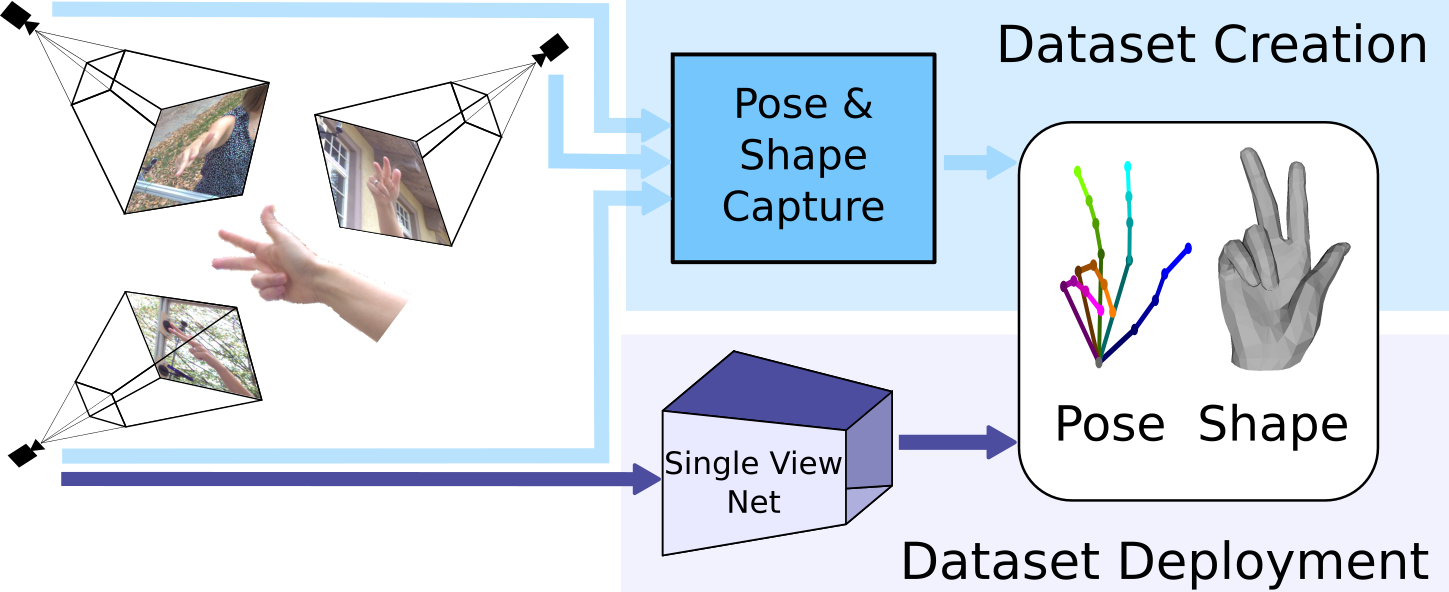}
\caption{We create a hand dataset via a novel iterative procedure that utilizes multiple views and sparse annotation followed by verification. This results in a large scale real world data set with pose and shape labels, which can be used to train single-view networks that have superior cross-dataset generalization performance on pose and shape estimation.}
\label{fig:teaser}
\vspace{-1em}
\end{figure}

Synthetic datasets use deformable hand models with texture information and render this model under varying pose configurations. As with all rendered datasets, it is difficult to model the wide set of characteristics of real images, such as varying illumination, camera lens distortion, motion blur, depth of field and debayering. Even more importantly, rendering of hands requires samples from the true distribution of feasible and realistic hand poses. In contrast to human pose, such distributional data does not exist to the same extent. Consequently, synthetic datasets are either limited in the variety of poses or sample many unrealistic poses.

\begin{figure*}[!t] 
\centering
\begin{minipage}{0.95\linewidth}
\begin{tabular}{@{}c@{}c@{}c@{}@{}c@{}c@{}c@{}}
 & Training Set &  &  & Evaluation Set & \\
 \includegraphics[width=.16\linewidth]{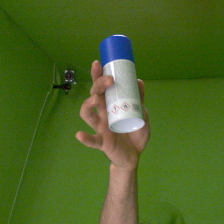}& 
 \includegraphics[width=.16\linewidth]{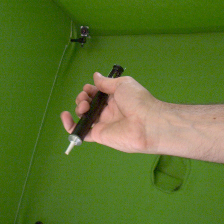}&
 \includegraphics[width=.16\linewidth]{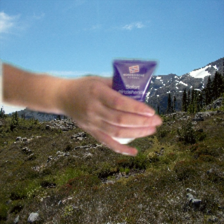}&
 
 \hspace{.014\linewidth}
 
 \includegraphics[width=.16\linewidth]{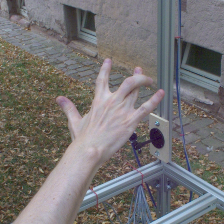}&
 \includegraphics[width=.16\linewidth]{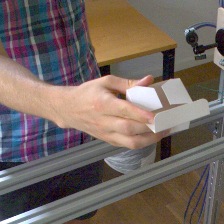}&
 \includegraphics[width=.16\linewidth]{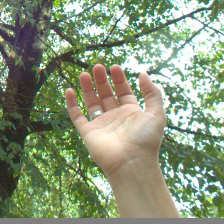}\\ [-\dp\strutbox]
 
 \includegraphics[width=.16\linewidth]{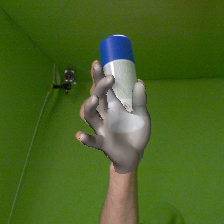}&
 \includegraphics[width=.16\linewidth]{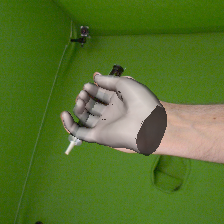}&
 \includegraphics[width=.16\linewidth]{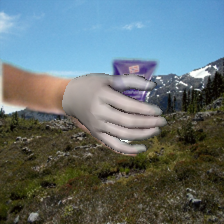} &
 
 \hspace{.014\linewidth}
     
 \includegraphics[width=.16\linewidth]{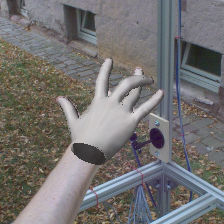}&
 \includegraphics[width=.16\linewidth]{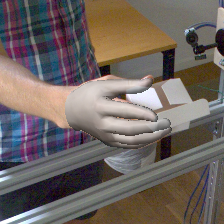}&
 \includegraphics[width=.16\linewidth]{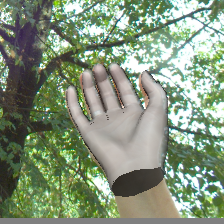}\\
\end{tabular}
\end{minipage}
\vspace{-.8em}
\caption{Examples from our proposed dataset showing images (top row) and hand shape annotations (bottom row). The training set contains composited images from green screen recordings, whereas the evaluation set contains images recorded indoors and outdoors. The dataset features several subjects as well as object interactions.
}
\label{fig:dataset_examples}
\vspace{-1em}
\end{figure*}

Capturing a dataset of real human hands requires annotation in a post-processing stage. 
In single images, manual annotation is difficult and cannot be easily crowd sourced due to occlusions and ambiguities. Moreover, collecting and annotating a large scale dataset is a respectable effort. 

In this paper, we analyze how these limitations affect the ability of single-view hand pose estimation to generalize across datasets and to \emph{in-the-wild} real application scenarios. We find that datasets show excellent performance on the respective evaluation split, but have rather poor performance on other datasets, i.e., we see a classical dataset bias. 

As a remedy to the dataset bias problem, we created a new large-scale dataset by increasing variation between samples. We collect a real-world dataset and develop a methodology that allows us to automate large parts of the labeling procedure, while manually ensuring very high-fidelity annotations of 3D pose and 3D hand shape. One of the key aspects is that we record synchronized images from multiple views, an idea already used previously in \cite{ballan_motion_2012, simon2017hand}. The multiple views remove many ambiguities and ease both the manual annotation and automated fitting. The second key aspect of our approach is a semi-automated \emph{human-in-the-loop} labeling procedure with a strong bootstrapping component. Starting from a sparse set of 2D keypoint annotations (e.g., finger tip annotations) and semi-automatically generated segmentation masks, we propose a hand fitting method that fits a deformable hand model~\cite{romero2017embodied} to a set of multi-view input. This fitting yields both 3D hand pose and shape annotation for each view. We then train a multi-view 3D hand pose estimation network using these annotations. This network predicts the 3D hand pose for unlabeled samples in our dataset along with a confidence measure. By verifying confident predictions and annotating least-confident samples in an iterative procedure, we acquire $11592$ annotations with moderate manual effort by a human annotator. 

The dataset spans $32$ different people and features fully articulated hand shapes, a high variation in hand poses and also includes interaction with objects. Part of the dataset, which we mark as training set, is captured against a green screen. Thus, samples can easily be composed with varying background images. The test set consists of recordings in different indoor and outdoor environments; see Figure~\ref{fig:dataset_examples} for sample images and the corresponding annotation.  

Training on this dataset clearly improves cross-dataset generalization compared to training on existing datasets. Moreover, we are able to train a network for full 3D hand shape estimation from a single RGB image. For this task, there is not yet any publicly available data, neither for training nor for benchmarking. Our dataset is available on our project page and therefore can serve both as training and benchmarking dataset for future research in this field.

\definecolor{darkblue}{rgb}{0.0, 0.0, 0.55}
\newcommand{\fr}[1]{$\mathbf{#1}$}  
\newcommand{\se}[1]{\textcolor{blue}{$#1$}} 
\newcommand{\tr}[1]{\textcolor{cyan}{$#1$}} 
\begin{table*}[htb!]
\begin{center}
\begin{tabular}{|c|c|c|c|c|c|c|c|c||c|}
\hline
\diagbox[width=5em]{train}{eval} & STB & RHD & GAN & PAN & LSMV & FPA & HO-3D & \textbf{Ours} & \thead{Average\\Rank}\\
\hline\hline
STB \cite{zhang_3d_2016} &\fr{0.783}& $0.179$  & $0.067$  & $0.141$  & $0.072$  & $0.061$  & $0.138$ & $0.138$ & $6.0$\\
RHD \cite{zb2017hand} & $0.362$ &\fr{0.767}&\tr{0.184}&\tr{0.463}&\se{0.544}& $0.101$  & \se{0.450} &\se{$0.508$} & $2.9$\\
GAN \cite{mueller2018ganerated} & $0.110$  & $0.103$  &\fr{0.765}& $0.092$  & $0.206$  & \tr{0.180}  & $0.087$ &  $0.183$ & $5.4$\\
PAN \cite{joo2018total} & \tr{0.459}  &\tr{0.316}& $0.136$  &\fr{0.870}& $0.320$  &\se{0.184}& \tr{0.351} & \tr{0.407} & $3.0$\\
LSMV \cite{gomez2019large} & $0.086$  & $0.209$  & $0.152$  & $0.189$  &\fr{0.717}& $0.129$  & $0.251$ & $0.276$ & $4.1$\\
FPA \cite{garcia2018first} & $0.119$  & $0.095$  & $0.084$  & $0.120$  & $0.118$  &\fr{0.777}& $0.106$ & $0.163$ & $6.0$\\
HO-3D \cite{hampali2019ho} & $0.154$  & $0.130$  & $0.091$  & $0.111$  & $0.149$  & $0.073$ & - & $0.169$ & $6.1$ \\
\textbf{Ours}&\se{0.473}&\se{0.518}&\se{0.217}&\se{0.562}&\tr{0.537}& $0.128$ & \fr{0.557}  &\fr{0.678} & $2.2$\\
\hline
\end{tabular}
\vspace{-.2em}
\caption{
This table shows cross-dataset generalization measured as area under the curve (AUC) of percentage of correct keypoints following \cite{zb2017hand}. 
Each row represents the training set used and each column the evaluation set. The last column shows the average rank each training set achieved across the different evaluation sets.
The top-three ranking training sets for each evaluation set are marked as follows: \textbf{first}, \textcolor{blue}{second} or \textcolor{cyan}{third}. Note that the evaluation set of \textit{HO-3D} was not available at time of submission, therefore one table entry is missing and the other entries within the respective column report numbers calculated on the training set.
}\label{tab:cross_dataset_gen}
\end{center}
\vspace{-2em}
\end{table*}

\section{Related Work}\label{sec:related_work}
Since datasets are crucial for the success of 3D hand pose and shape estimation, there has been much effort on acquiring such data.

In the context of hand shape estimation, the majority of methods fall into the category of model-based techniques. These approaches were developed in a strictly controlled environment and utilize either depth data directly \cite{tkach2016sphere, tkach2017online, tzionas2014capturing} or use multi-view stereo methods for reconstruction \cite{ballan_motion_2012}. More related to our work are approaches that fit statistical human shape models to observations \cite{bogo2016keep, Lassner_2017_CVPR} from \emph{in-the-wild} color images as input. Such methods require semi-automatic methods to acquire annotations such as keypoints or segmentation masks for each input image to guide the fitting process.

Historically, acquisition methods often incorporated markers onto the hand that allow for an easy way to estimate pose. Common choices are infrared markers~\cite{hillebrand2006inverse}, color coded gloves~\cite{wang2009real}, or electrical sensing equipment~\cite{zimmerman1987hand}. This alters hand appearance and, hence, makes the data less valuable for training discriminative methods.

Annotations can also be provided manually on hand images~\cite{OccludedHands_ICCV2017, RealtimeHO_ECCV2016, zhang_3d_2016}. However, the annotation is limited to visible regions of the hand. Thus, either the subject is required to retain from complex hand poses that result in severe self-occlusions, or only a subset of hand joints can be annotated.

To avoid occlusions and annotate data at larger scale, Simon \etalcite{simon2017hand} leveraged a multi-view recording setup. They proposed an iterative bootstrapping approach to detect hand keypoints in each view and triangulate them to generate 3D point hypotheses. While the spirit of our data collection strategy is similar, we directly incorporate the multi-view information into a neural network for predicting 3D keypoints and our dataset consists of both pose and shape annotations.

Since capturing real data comes with an expensive annotation setup and process, more methods rather deployed synthetic datasets recently~\cite{OccludedHands_ICCV2017, zb2017hand}.

\section{Analysis of Existing Datasets} \label{sec:sota_datasets}
We thoroughly analyze state-of-the-art datasets used for 3D hand pose estimation from single RGB images by testing their ability to generalize to unseen data. 
We identify seven state-of-the-art datasets that provide samples in the form of an RGB image and the accompanying 3D keypoint information as shown in \tabref{tab:dataset_overview}.

\subsection{Considered Datasets}

\textbf{Stereo Tracking Benchmark} (\textbf{STB}) \cite{zhang_3d_2016} dataset is one of the first and most commonly used datasets to report performance of 3D keypoint estimation from a single RGB image. The annotations are acquired manually limiting the setup to hand poses where most regions of the hands are visible. Thus, the dataset shows a unique subject posing in a frontal pose with different background scenarios and without objects. 

The \textbf{Panoptic} (\textbf{PAN}) dataset \cite{joo2018total} was created using a dense multi-view capture setup consisting of 10 RGB-D sensors, 480 VGA and 31 HD cameras. It shows humans performing different tasks and interacting with each other. There are 83 sequences publicy available and 12 of them have hand annotation. We select \textit{171204\_pose3} to serve as evaluation set and use the remaining $11$ sequences from the \emph{range motion, haggling and tools} categories for training.

Garcia \etal\cite{garcia2018first} proposed the \textbf{First-person hand action benchmark} (\textbf{FPA}), a large dataset that is recorded from an egocentric perspective and annotated using magnetic sensors attached to the finger tips of the subjects. Wires run along the fingers of the subject altering the appearance of the hands significantly. 6 DOF sensor measurements are utilized in an inverse kinematics optimization of a given hand model to acquire the full hand pose annotations.

Using the commercial Leap Motion device \cite{leapmotion} for keypoint annotation, Gomez \etal\cite{gomez2019large} proposed the \textbf{Large-scale Multiview 3D Hand Pose Dataset} (\textbf{LSMV}). Annotations given by the device are transformed into $4$ calibrated cameras that are approximately time synchronized. Due to the limitations of the sensor device, this dataset does not show any hand-object interactions. 

The \textbf{Rendered Hand Pose Dataset} (\textbf{RHD}) proposed by Zimmermann \etal\cite{zb2017hand} is a synthetic dataset rendered from $20$ characters performing $31$ different actions in front of a random background image without hand object interaction. 

Building on the SynthHands \cite{OccludedHands_ICCV2017} dataset Mueller \etal\cite{mueller2018ganerated} presented the \textbf{GANerated} (\textbf{GAN}) dataset. SynthHands was created by retargeting measured human hand articulation to a rigged meshed model in a mixed reality approach. This allowed for hand object interaction to some extend, because the subject could see the rendered scene in real time and pose the hand accordingly. In the following \textit{GANerated} hand dataset, a CycleGAN approach is used to bridge the synthetic to real domain shift. 

Recently, Hampali \etal\cite{hampali2019ho} proposed an algorithm for dataset creation deploying an elaborate optimization scheme incorporating temporal and physical consistencies, as well as silhouette and depth information. The resulting dataset is referred to as \textbf{HO-3D}.

\begin{table}[tb]
\footnotesize
\begin{center}
\resizebox{\columnwidth}{!}{
\begin{tabular}{|c|c|c|c|c|c|c|}
\hline
dataset & num. &   num.      & real & obj- & shape & labels\\
        & frames& subjects &  & ects & & \\
\hline\hline
STB \cite{zhang_3d_2016} & \SI{15}{k}~/~\SI{3}{k} & $1$ & \Y & \N & \N & manual \\
PAN \cite{zb2017hand} & \SI{641}{k}~/~\SI{34}{k} & $>10$ & \Y & \Y & \N & MVBS \cite{simon2017hand} \\
FPA \cite{garcia2018first} & \SI{52}{k}~/~\SI{53}{k} & $6$ & \Y & \Y & \N & marker \\
LSMV \cite{gomez2019large} & \SI{117}{k}~/~\SI{31}{k} & $21$ & \Y & \N & \N & leapmotion \\
RHD \cite{zb2017hand} & \SI{41}{k}~/~\SI{2.7}{k} & 20 & \N & \N & \N & synthetic \\
GAN \cite{mueller2018ganerated} & \SI{266}{k}~/~\SI{66}{k} & - & \N & \Y & \N & synthetic \\
HO-3D \cite{hampali2019ho} & \SI{11}{k}~/ - & 3 & \Y & \Y & \Y & automatic \cite{hampali2019ho} \\
Ours & \SI{33}{k}~/~\SI{4}{k} & 32 & \Y & \Y & \Y & hybrid \\
\hline
\end{tabular}
}
\caption{State-of-the-art datasets for the task of 3D keypoint estimation from a single color image used in our analysis. We report dataset size in number of frames, number of subjects, if it is real or rendered data, regarding hand object interaction, if shape annotation is provided and which method was used for label generation.
}\label{tab:dataset_overview}
\end{center}
\vspace{-2.5em}
\end{table}

\subsection{Evaluation Setup} 
We trained a state-of-the-art network architecture~\cite{iqbal2018hand} that takes as input an RGB image and predicts 3D keypoints on the training split of each of the datasets and report its performance on the evaluation split of all other datasets. For each dataset, we either use the standard training/evaluation split reported by the authors or create an $80\%/20\%$ split otherwise; see the supplementary material for more details.

The single-view network takes an RGB image $\mathbf{I}$ as input and infers 3D hand pose $\mathbf{P} = \{ \mathbf{p}_k \}$ with each $\mathbf{p}_k \in \mathbb{R}^3$, representing a predefined landmark or keypoint situated on the kinematic skeleton of a human hand. Due to scale ambiguity, the problem to estimate real world 3D keypoint coordinates in a camera centered coordinate frame is ill-posed. Hence, we adopt the problem formulation of \cite{iqbal2018hand} to estimate coordinates in a root relative and scale normalized fashion:
\begin{align}
    \mathbf{p}_k = s \cdot \hat{\mathbf{p}}_k = s \cdot 
    \begin{pmatrix} \hat{x}_k\\ \hat{y}_k\\ \hat{z}_k \end{pmatrix} =
    \begin{pmatrix} \hat{x}_k\\ \hat{y}_k\\ \hat{z}^\text{rel}_k + \hat{z}^\text{root} \end{pmatrix},
\end{align}
where the normalization factor $s$ is chosen as the length of one reference bone in the hand skeleton, $\hat{z}^\text{root}$ is the root depth and $\hat{z}^\text{rel}_k$ the relative depth of keypoint $k$. We define the resulting 2.5D representation as:

\begin{equation}
    \hat{\mathbf{p}}_{\text{rel}_k} = \begin{pmatrix} \hat{x}_k,~ \hat{y}_k,~ \hat{z}^\text{rel}_k \end{pmatrix}^\text{T}.
\end{equation}
Given scale constraints and 2D projections of the points in a calibrated camera, 3D hand pose $\mathbf{P}$ can be recovered from $\hat{\mathbf{P}}_\text{rel}$. For details about this procedure we refer to \cite{iqbal2018hand}.

We train the single-view network using the same hyperparameter choices as Iqbal \etal~\cite{iqbal2018hand}. However, we use only a single stage and reduce the number of channels in the network layers, which leads to a significant speedup in terms of training time at only a marginal decrease in accuracy. We apply standard choices of data augmentation including color, scale and translation augmentation as well as rotation around the optical axis. We apply this augmentation to each of the datasets.  

\subsection{Results}
It is expected that the network performs the best on the dataset it was trained on, yet it should also provide reasonable predictions for unseen data when being trained on a dataset with sufficient variation (e.g., hand pose, viewpoint, shape, existence of objects, etc.).

\tabref{tab:cross_dataset_gen} shows for each existing training dataset the network is able to generalize to the respective evaluation split and reaches the best results there. On the other hand, performance drops substantially when the network is tested on other datasets. 

Both \textit{GAN} and \textit{FPA} datasets appear to be especially hard to generalize indicating that their data distribution is significantly different from the other datasets. For \textit{FPA} this stems from the appearance change due to the markers used for annotation purposes. The altered appearance gives the network trained on this dataset strong cues to solve the task that are not present for other datasets at evaluation time. Thus, the network trained on \textit{FPA} performs poorly when tested on other datasets. Based on visual inspection of the \textit{GAN} dataset, we hypothesize that subtle changes like missing hand texture and different color distribution are the main reasons for generalization problems. 
We also observe that while the network trained on \text{STB} does not perform well on remaining datasets, the networks trained on other datasets show reasonable performance on the evaluation split of \textit{STB}. We conclude that a good performance on \textit{STB} is not a reliable measure for how a method generalizes to unseen data.

Based on the performance of each network, we compute a cumulative ranking score for each dataset that we report in the last column of Table~\ref{tab:cross_dataset_gen}. To calculate the cumulative rank we assign ranks for each column of the table separately according to the performance the respective training sets achieve. The cumulative rank is then calculated as average over all evaluation sets, \ie rows of the table. Based on these observations, we conclude that there is a need for a new benchmarking dataset that can provide superior generalization capability. 

We present the \ourdb~Dataset to archieve this goal. It consists of real images, provides sufficient viewpoint and hand pose variation, and shows samples both with and without object interactions. Consequently, the single-view network trained on this dataset achieves a substantial improvement in terms of ranking for cross-dataset generalization. We next describe how we acquired and annotated this dataset.

\begin{figure}[htb!]
\centering
\includegraphics[width=1.0\columnwidth]{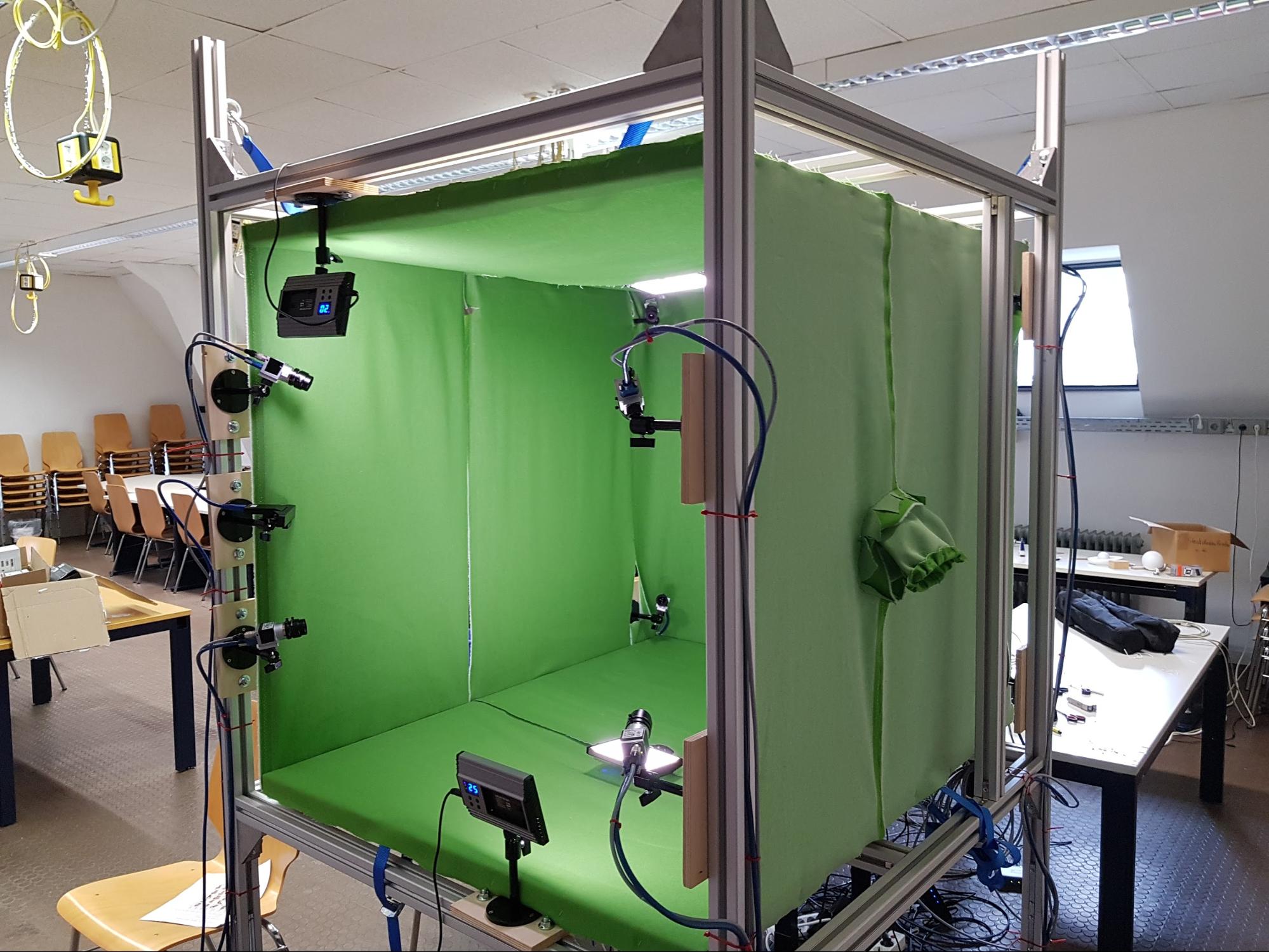}
\caption{Recording setup with $8$ calibrated and temporally synchronized RGB cameras located at the corners of a cube. A green screen background can be mounted into the the setup, enabling easier background subtraction.}
\label{fig:setup}
\vspace{-1.2em}
\end{figure}

\section{\ourdb~Dataset} \label{sec:freihand}
The dataset was captured with the multi-view setup shown in \figref{fig:setup}. 
The setup is portable enabling both indoor and outdoor capture. We capture hand poses from $32$ subjects of different genders and ethnic backgrounds. Each subject is asked to perform actions with and without objects.
To capture hand-object interactions, subjects are given a number of everyday household items that allow for reasonable one-handed manipulation and are asked to demonstrate different grasping techniques. More information is provided in the supplementary material. 

To preserve the realistic appearance of hands, no markers are used during the capture. Instead we resort to post-processing methods that generate 3D labels. Manual acquisition of 3D annotations is obviously unfeasible. An alternative strategy is to acquire 2D keypoint annotations for each input view and utilize the multi-view camera setup to lift such annotations to 3D similar to Simon~\etal~\cite{simon2017hand}. 

We found after initial experiments that current 2D hand pose estimation methods perform poorly, especially in case of challenging hand poses with self- and object occlusions. Manually annotating all 2D keypoints for each view is prohibitively expensive for large-scale data collection. Annotating all 21 keypoints across multiple-views with a specialized tool takes about 15 minutes for each multi-view set. Furthermore, keypoint annotation alone is not sufficient to obtain shape information. 

\begin{figure}
\centering
    \includegraphics[width=.6\columnwidth, height=.6\columnwidth]{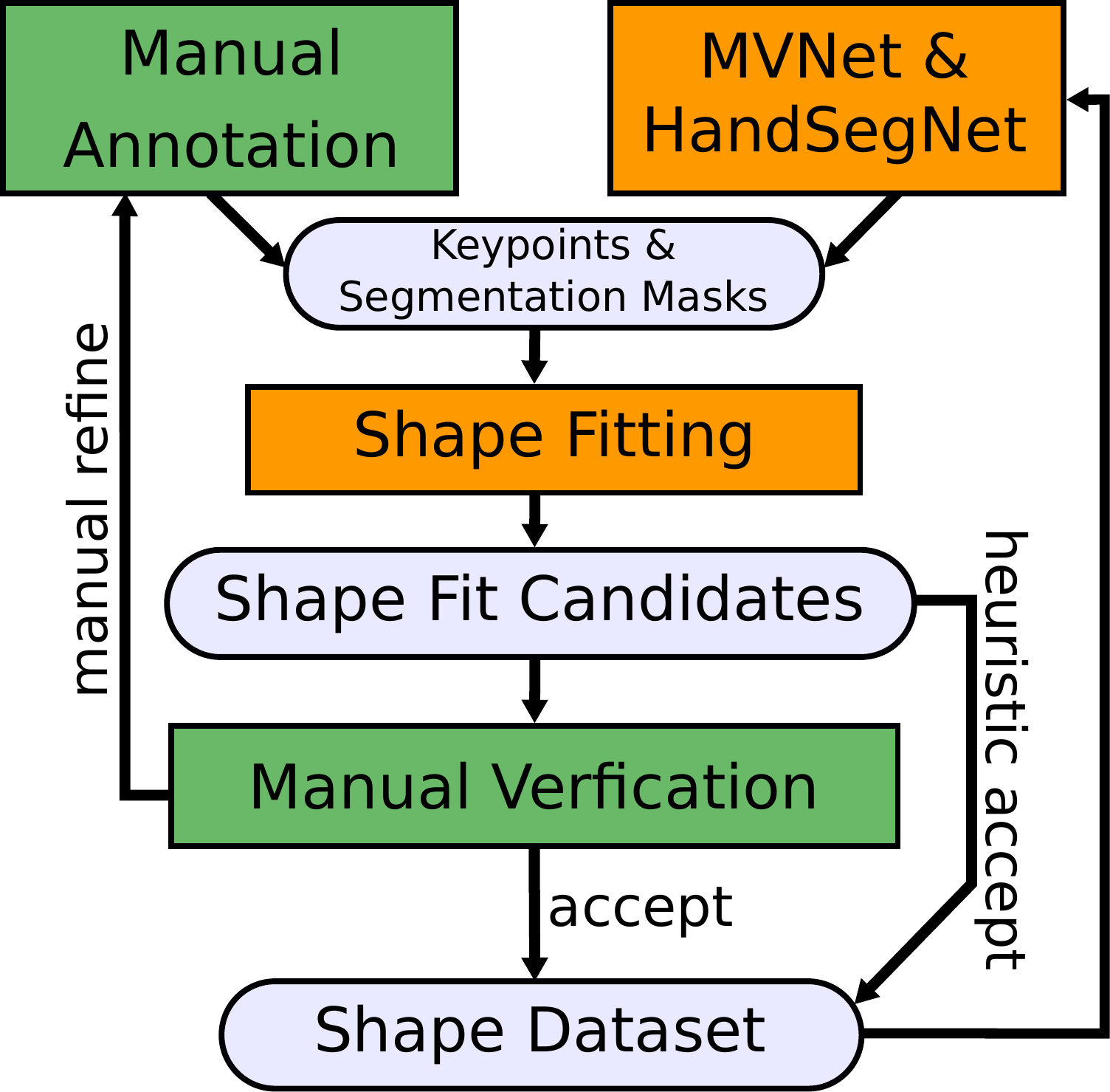}
\caption{The dataset labeling workflow starts from manual annotation followed by the shape fitting process described in \ref{sec:fitting}, which yields candidate shape fits for our data samples. Sample fits are manually verified allowing them to be accepted, rejected or queued for further annotation. Alternatively a heuristic can accept samples without human interaction. The initial dataset allows for training the networks involved, which for subsequent iterations of the procedure, can predict information needed for fitting. The labeling process can be bootstrapped, allowing more accepted samples to accumulate in the dataset.}\label{fig:bootstrap}
\vspace{-1.2em}
\end{figure}

We address this problem with a novel bootstrapping procedure (see \figref{fig:bootstrap}) composed of a set of automatic methods that utilize sparse 2D annotations.
Since our data is captured against a green screen, the foreground can be extracted automatically. Refinement is needed only to co-align the segmentation mask with the hand model's wrist.
In addition, a sparse set of six 2D keypoints (finger tips and wrist) is manually annotated.
These annotations are relatively cheap to acquire at a reasonably high quality. For example, manually correcting a segmentation mask takes on average 12 seconds, whereas annotating a keypoint takes around 2 seconds.
Utilizing this information we fit a deformable hand model to multi-view images using a novel fitting process described in Section~\ref{sec:fitting}. This yields candidates for both 3D hand pose and shape labels.
These candidates are then manually verified, before being added to a set of labels.

Given an initial set of labels, we train our proposed network, \textit{MVNet}, that takes as inputs multi-view images and predicts 3D keypoint locations along with a confidence score, described in Section~\ref{sec:MVNet}. Keypoint predictions can be used in lieu of manually annotated keypoints as input for the fitting process. This bootstrapping procedure is iterated. 
The least-confident samples are manually annotated (Section~\ref{sec:refinement}). With this \emph{human-in-the-loop} process, we quickly obtain a large scale annotated dataset. Next we describe each stage of this procedure in detail.  

\subsection{Hand Model Fitting with Sparse Annotations}\label{sec:fitting}

Our goal is to fit a deformable hand shape model to observations from multiple views acquired at the same time.  We build on the statistical \textit{MANO} model, proposed by Romero~\etal\cite{romero2017embodied}, which is parameterized by $\bm{\theta} \in \mathbb{R}^{61}$. The model parameters $\bm{\theta} = (\bm{\alpha},  \bm{\beta}, \bm{\gamma})^\text{T}$ include shape $\bm{\alpha} \in \mathbb{R}^{10}$, articulation $\bm{\beta} \in \mathbb{R}^{45}$ as well as global translation and orientation $\bm{\gamma} \in \mathbb{R}^6$.
Using keypoint and segmentation information we optimize a multi-term loss, 
\begin{align}
    \mathcal{L} = \mathcal{L}^\text{2D}_\text{kp} + \mathcal{L}^\text{3D}_\text{kp} + \mathcal{L}_\text{seg} + \mathcal{L}_\text{shape} + \mathcal{L}_\text{pose} \text{,}
    \label{eq:optim_obj_all}
\end{align}
to estimate the model parameters $\bm{\tilde{\theta}}$, where the tilde indicates variables that are being optimized. We describe each of the terms in \eqref{eq:optim_obj_all} next.

\noindent
\textbf{2D Keypoint Loss $\mathcal{L}^\text{2D}_\text{kp}$: } 
The loss is the sum of distances between the 2D projection $\Pi^i$ of the models' 3D keypoints ${\bm{\tilde{p}}_k} \in \mathbb{R}^3$ to the 2D annotations $\bm{q}_k^{i}$ over views $i$ and visible keypoints $k \in V_i$:
\begin{equation}
    \mathcal{L}^\text{2D}_\text{kp} = w^\text{2D}_\text{kp}\cdot \sum_i \sum_{k \in V_i} \cdot \norm{ \bm{q}_k^{i} - \Pi^i(\bm{\tilde{p}}_k)} \text{.}
    \label{eq:optim_obj_annotation}
\end{equation}

\noindent
\textbf{3D keypoint Loss $\mathcal{L^\text{3D}_\text{kp}}$:} This loss is defined in a similar manner as \eqref{eq:optim_obj_annotation}, but over 3D keypoints. Here, $\bm{p}_k$ denotes the 3D keypoint annotations, whenever such annotations are available (e.g., if predicted by \textit{MVNet}), 
\begin{equation}
    \mathcal{L}^\text{3D}_\text{kp} = w^\text{3D}_\text{kp} \cdot \sum_i \sum_{k \in V_i} \norm{ \bm{p}_k - \bm{\tilde{p}}_k} \text{.}
    \label{eq:optim_obj_3d}
\end{equation}

\noindent
\textbf{Segmentation Loss $\mathcal{L_{\text{seg}}}$:} For shape optimization we use a sum of $l_2$ losses between the model dependent mask $\tilde{\bm{M}}^i$ and the manual annotation $\bm{M}^i$ over views $i$:

\begin{equation}
    \mathcal{L_{\text{seg}}} = w_\text{seg} \cdot \sum_i (\norm{ \bm{M}^i - \bm{\tilde{M}}^i} + \norm{ \text{EDT}( \bm{M}^i ) \cdot \bm{\Tilde{M}}^i}).
    \label{eq:optim_obj_seg}
\end{equation}

\noindent
Additionally, we apply a silhouette term based on the Euclidean Distance Transform (EDT). Specifically, we apply a symmetric EDT to $\bm{M}^i$, which contains the distance to the closest boundary pixel at every location.

\noindent
\textbf{Shape Prior $\mathcal{L}_{\text{shape}}$:} For shape regularization we employ
\begin{equation}
    \mathcal{L_{\text{shape}}} = w_\text{shape} \cdot \norm{\bm{\tilde{\beta}}},
    \label{eq:optim_obj_shape}
\end{equation}
which enforces the predicted shape to stay close to the mean shape of \textit{MANO}. 

\noindent
\textbf{Pose Prior $\mathcal{L_{\text{pose}}}$:} The pose prior has two terms. The first term applies a regularization on the PCA coefficients $a_j$ used to represent the pose $\bm{\tilde{\alpha}}$ in terms of PCA basis vectors $\bm{c}_j$ (i.e., $\bm{\tilde{\alpha}} = \sum_j \tilde{a}_j \cdot \bm{c}_{j}$). This regularization enforces predicted poses to stay close to likely poses with respect to the PCA pose space of \textit{MANO}.
The second term regularizes the distance of the current pose $\bm{\tilde{\alpha}}$, to the $N$ nearest neighbors of a hand pose dataset acquired from \cite{garcia2018first}:
\begin{align}
    \mathcal{L_{\text{pose}}} = w_\text{pose} \cdot \sum_{j} \norm{\tilde{a}_j}  + w_\text{nn} \cdot \sum_{n \in N} \norm{\bm{\alpha}^n - \bm{\tilde{\alpha}}} \text{.}
    \label{eq:optim_obj_pose}
\end{align}

We implement the fitting process in Tensorflow~\cite{abadi2016tensorflow} and use \textit{MANO} to implement a differentiable mapping from $\tilde{\bm{\theta}}$ to 3D model keypoints $\tilde{\bm{p}}_k$ and 3D model vertex locations $\tilde{\bm{V}} \in \mathbb{R}^{778 \times 3}$. We adopt the Neural Renderer~\cite{kato2018neural} to render the segmentation masks $\bm{\tilde{M}}^i$ from the hand model vertices $\tilde{\bm{V}}$ and use the ADAM optimizer \cite{adam_kingsma} to minimize:
\begin{align}
    \bm{\theta} = \argmin_{\tilde{\theta}} ( \mathcal{L(\bm{\tilde{\theta}})} )
    \label{eq:optim_minim}
\end{align}

\begin{figure}[ht]
\centering
    \includegraphics[width=.9\columnwidth]{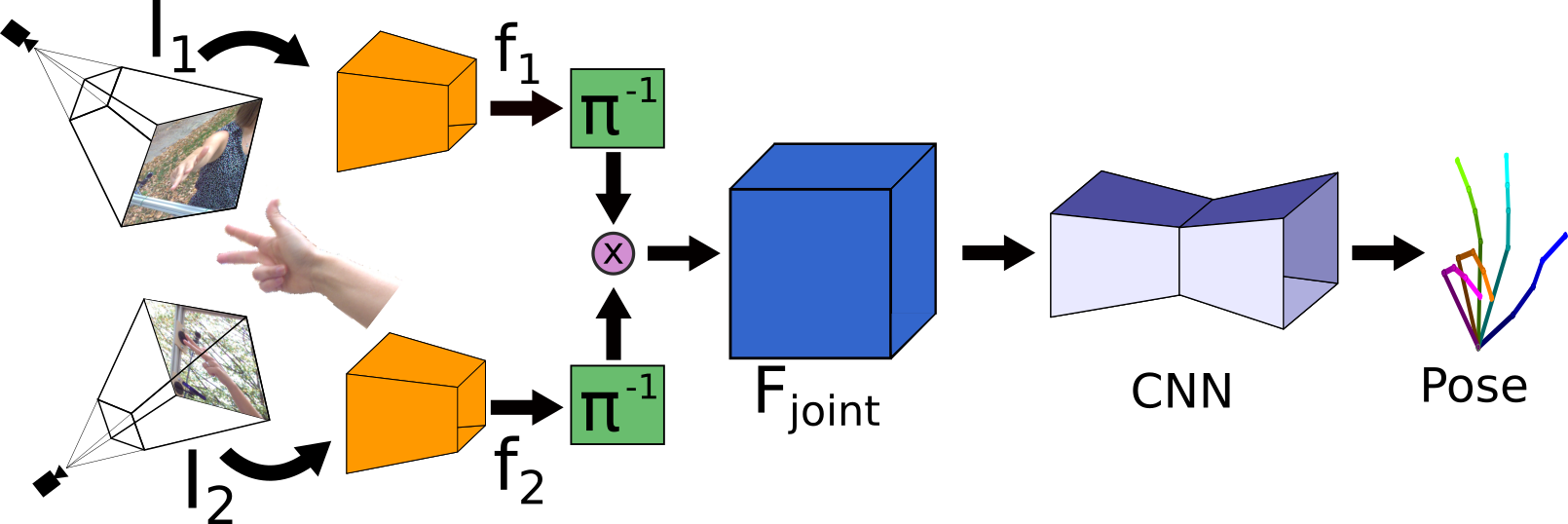}
\caption{\textit{MVNet} predicts a single hand pose $\bm{P}$ using images of all $8$ views (for simplicity only $2$ are shown). Each image is processed separately by a 2D CNN that is shared across views. This yields 2D feature maps $\bm{f}_i$. These are individually reprojected into a common coordinate frame using the known camera calibration to obtain $\bm{F}_i=\Pi^{-1}(\bm{f}_i)$. The $\bm{F}_i$ are aggregated over all views and finally a 3D CNN localizes the 3D keypoints within a voxel representation.}\label{fig:MVNet}
\vspace{-1em}
\end{figure}

\subsection{MVNet: Multiview 3D Keypoint Estimation} \label{sec:MVNet}

To automate the fitting process, we seek to estimate 3D keypoints automatically. We propose \textit{MVNet} shown in \figref{fig:MVNet} that aggregates information from all eight camera images $\bm{I}_i$
and predicts a single hand pose  $\bm{P} = \{\bm{p}_k\}$.
We use a differentiable unprojection operation, similar to Kar~\etalcite{kar2017learning}, to aggregate features from each view into a common 3D volume.

To this end, we formulate the keypoint estimation problem as a voxel-wise regression task:
\begin{align}
    \mathcal{L}_{\text{MVNet}} =\frac{1}{K} \sum_k \norm{\bm{S}_k - \bm{\tilde{S}}_k},
    \label{eq:MVNet_obj}
\end{align}
where $\bm{\tilde{S}}_k \in \mathbb{R}^{N \times N \times N}$ represents the prediction of the network for keypoint $k$ and $\bm{S}_k$ is the ground truth estimate we calculate from validated \textit{MANO} fits. $\bm{S}_k$ is defined as a normalized Gaussian distribution centered at the true keypoint location. The predicted point $\bm{\tilde{p}}_k$ is extracted as maximal location in $\bm{\tilde{S}}_k$. Furthermore, we define the confidence $c$ of a prediction as maximum along the spatial and average over the keypoint dimension:
\begin{align}
    c = \frac{1}{K} \sum_k ( \max_{i, j, l} \bm{\tilde{S}}_k(i, j, l) ) .
    \label{eq:MVNet_conf}
\end{align}
Additional information can be found in the supplemental material.

\subsection{Iterative Refinement}\label{sec:refinement}
In order to generate annotations at large scale, we propose an iterative, \emph{human-in-the-loop} procedure which is visualized in \figref{fig:bootstrap}. For initial bootstrapping we use a set of manual annotations to generate the initial dataset $\mathcal{D}_0$. In iteration $i$ we use dataset $\mathcal{D}_i$, a set of images and the corresponding \textit{MANO} fits, to train \textit{MVNet} and \textit{HandSegNet} \cite{zb2017hand}. \textit{MVNet} makes 3D keypoint predictions along with confidence scores for the remaining unlabeled data and \textit{HandSegNet} predicts hand segmentation masks. Using these predictions, we perform the hand shape fitting process of Section \ref{sec:fitting}. Subsequently, we perform \emph{verification} that either accepts, rejects or partially annotates some of these data samples.

\begin{table}
\begin{center}
\resizebox{\columnwidth}{!}{
\begin{tabular}{|c|c|c|c|}
\hline
Method & mesh error $\downarrow$ & F@5mm $\uparrow$ & F@15mm $\uparrow$\\
\hline\hline
Mean shape & $1.63$ & $0.340$ & $0.839$ \\
MANO Fit \cmt{run329}& $1.44$ & $0.416$ & $0.880$ \\
    MANO CNN \cmt{run339}& $\mathbf{1.07}$ & $\mathbf{0.529}$ & $\mathbf{0.935}$ \\
Boukhayma \etal\cite{boukhayma20193d} & $1.30$ & $0.435$ & $0.898$ \\
Hasson \etal\cite{hasson2019learning} & $1.32$ & $0.436$ & $0.908$ \\
\hline
\end{tabular}
} 
\caption{This table shows shape prediction performance on the evaluation split of \ourdb~ after alignment. We report two measures: The mean mesh error and the F-score at two different distance thresholds.}\label{tab:shape_results}
\end{center}
\vspace{-2.2em}
\end{table}

\textbf{Heuristic Verification.} We define a heuristic consisting of three criteria to identify data samples with good \textit{MANO} fits. First, we require the mean \textit{MVNet} confidence score to be above $0.8$ and all individual keypoint confidences to be at least $0.6$, which enforces a minimum level of certainty on the 3D keypoint prediction. Second, we define a minimum threshold for the intersection over union (IoU) between predicted segmentation mask and the mask derived from the \textit{MANO} fitting result. We set this threshold to be $0.7$ on average across all views while also rejecting samples that have more than $2$ views with an IoU below $0.5$. Third, we require the mean Euclidean distance between predicted 3D keypoints and the keypoints of the fitted MANO to be at most $0.5$ cm where no individual keypoint has a Euclidean distance greater than $1$ cm. We accept only samples that satisfy all three criteria and add these to the set $\mathcal{D}^{h}_i$.

\textbf{Manual Verification and Annotation.} The remaining unaccepted samples are sorted based on the confidence score of \textit{MVNet} and we select samples from the $50^{th}$ percentile upwards. We enforce a minimal temporal distance between samples selected to ensure diversity as well as choosing samples for which the current pose estimates are sufficiently different to a flat hand shape as measured by the Euclidean distance in the pose parameters. We ask the annotators to evaluate the quality of the \textit{MANO} fits for these samples. Any sample that is verified as a good fit is added to the set $\mathcal{D}^{m}_i$. For remaining samples, the annotator has the option of either discarding the sample
or provide additional annotations (e.g., annotating mislabeled finger tips) to help improve the fit. These additionally annotated samples are added to the set $\mathcal{D}^{l}_i$.

Joining the samples from all streams yields a larger labeled dataset
\begin{equation}
    \mathcal{D}_{i+1} = \mathcal{D}_{i} + \mathcal{D}^{h}_i + \mathcal{D}^{m}_i + \mathcal{D}^{l}_i
\end{equation}
which allows us to retrain both \textit{HandSegNet} and \textit{MVNet}. We repeated this process $4$ times to obtain our final dataset.

\section{Experiments}

\subsection{Cross-Dataset Generalization of \ourdb}
To evaluate the cross-dataset generalization capability of our dataset and to compare to the results of \tabref{tab:cross_dataset_gen}, we define the following training and evaluation split: there are samples with and without green screen and we chose to use all green screen recordings for training and the remainder for evaluation. Training and evaluation splits contain data from $24$ and $11$ subjects, respectively, with only $3$ subjects shared across splits. The evaluation split is captured in $2$ different indoor and $1$ outdoor location. We augmented the training set by leveraging the green screen for easy and effective background subtraction and creating composite images using new backgrounds. To avoid green color bleeding at the hand boundaries we applied the image harmonization method of Tsai \etal\cite{tsai2017deep} and the deep image colorization approach by Zhang \etal\cite{zhang2017real} separately to our data. Both the automatic and sampling variant of \cite{zhang2017real} were used. With the original samples this quadruples the training set size from \SI{33}{k} unique to \SI{132}{k} augmented samples. Examples of resulting images are shown in \figref{fig:dataset_examples}.

Given the training and evaluation split, we train the single view 3D pose estimation network on our data and test it across different datasets. As shown in \tabref{tab:cross_dataset_gen}, the network achieves strong accuracy across all datasets and ranks first in terms of cross-dataset generalization.

\subsection{3D Shape Estimation} \label{sec:shape_exp}

Having both pose and shape annotations, our acquired dataset can be used for training shape estimation models in a fully supervised way. In addition, it serves as the first real dataset that can be utilized for evaluating shape estimation methods. Building on the approach of Kanazawa~\etal\cite{kanazawa2018end}, we train a network that takes as input a single RGB image and predicts the \textit{MANO} parameters $\bm{\tilde{\theta}}$ using the following loss:
\begin{align}
    \mathcal{L} = w_\text{3D} &\norm{\bm{p}_k - \tilde{\bm{p}}_k} + w_\text{2D} \norm{\Pi (\bm{p}_k) - \Pi ( \tilde{\bm{p}} )} +   \nonumber \\
    w_\text{p} &\norm{\bm{\theta} - \tilde{\bm{\theta}}} \text{.}
\end{align}
We deploy $l_2$ losses for 2D and 3D keypoints as well as the model parameters and chose the weighting to $w_\text{3D} = 1000$, $w_\text{2D} = 10$ and $w_\text{p} = 1$.

We also provide two baseline methods, constant mean shape prediction, without accounting for articulation changes, and fits of the \textit{MANO} model to the 3D keypoints predicted by our single-view network. 

\begin{figure}[!htb] 
\centering
\begin{minipage}{0.85\linewidth}
\begin{tabular}{@{}c@{}c@{}c@{}}
 \includegraphics[width=.33\linewidth]{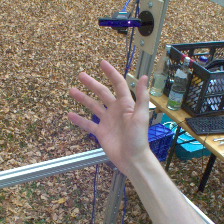}& 
 \includegraphics[width=.33\linewidth]{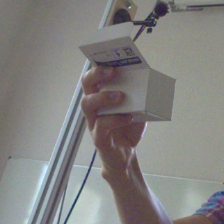}&
 \includegraphics[width=.33\linewidth]{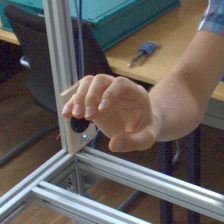}\\ [-\dp\strutbox]
 \includegraphics[width=.33\linewidth]{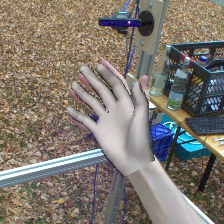}&
 \includegraphics[width=.33\linewidth]{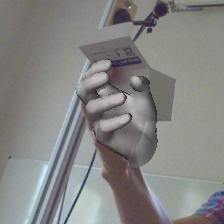}&
 \includegraphics[width=.33\linewidth]{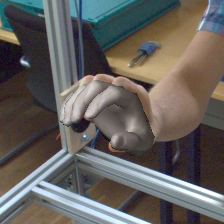}\\
 \includegraphics[width=.33\linewidth]{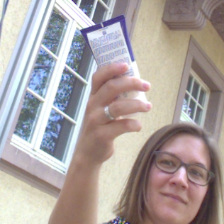}&
 \includegraphics[width=.33\linewidth]{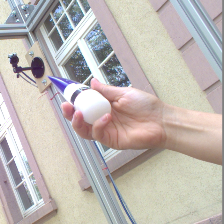}& 
 \includegraphics[width=.33\linewidth]{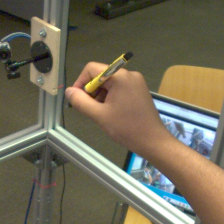}\\ [-\dp\strutbox]
 \includegraphics[width=.33\linewidth]{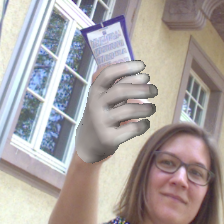}&
 \includegraphics[width=.33\linewidth]{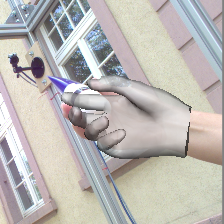}&
 \includegraphics[width=.33\linewidth]{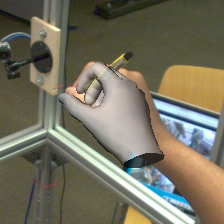}\\
\end{tabular}
\end{minipage}
\vspace{-.2em}
\caption{Given a single image (top rows), qualitative results of predicted hand shapes (bottom rows) are shown. Please note that we don't apply any alignment of the predictions with respect to the ground truth.}
\label{fig:examples_mano_pred}
\end{figure}

For comparison, we use two scores. The \emph{mesh error} measures the average Euclidean distance between corresponding mesh vertices in the ground truth and the predicted hand shape. 
We also evaluate the $F$-score \cite{knapitsch2017tanks} which, given a distance threshold, defines the harmonic mean between recall and precision between two sets of points \cite{knapitsch2017tanks}. In our evaluation, we use two distances: $F$@5mm and $F$@15mm to report the accuracy both at fine and coarse scale. In order to decouple shape evaluation from global rotation and translation, we first align the predicted meshes using Procrustes alignment. Results are summarized in \tabref{tab:shape_results}. Estimating \textit{MANO} parameters directly with a CNN performs better across all measures than the baseline methods.
The evaluation reveals that the difference in $F$-score is more pronounced in the high accuracy regime. Qualitative results of our network predictions are provided in \figref{fig:examples_mano_pred}.

\subsection{Evaluation of Iterative Labeling}
In the first step of iterative labeling process, we set $w^\text{2D}_\text{kp}=100$ and $w^\text{2D}_\text{kp}=0$ (since no 3D keypoint annotations are available), $w_{seg}= 10.0$, $w_\text{shape} = 100.0$, $w_\text{nn} = 10.0$, and $w_\text{pose} = 0.1$. (For subsequent iterations we set $w^\text{2D}_\text{kp}=50$ and $w^\text{3D}_\text{kp}=1000$.) Given the fitting results, we train \textit{MVNet} and test it on the remaining dataset. After the first verification step, $302$ samples are accepted. Validating a sample takes about $5$ seconds and we find that the global pose is captured correctly in most cases, but in order to obtain high quality ground truth, even fits with minor inaccuracies are discarded. 

We use the additional accepted samples to retrain \textit{MVNet} and \textit{HandSegNet} and iterate the process. At the end of the first iteration we are able to increase the dataset to $993$ samples, $140$ of which are automatically accepted by heuristic, and the remainder from verifying $1000$ samples. In the second iteration the total dataset size increases to $1449$, $289$ of which are automatically accepted and the remainder stems from verifying $500$ samples. In subsequent iterations the complete dataset size is increased to $2609$ and $4565$ samples, where heuristic accept yields $347$ and $210$ samples respectively. This is the dataset we use for the cross-dataset generalization (see \tabref{tab:cross_dataset_gen}) and shape estimation (see \tabref{tab:shape_results}) experiments.

We evaluate the effectiveness of the iterative labeling process by training a single view 3D keypoint estimation network on different iterations of our dataset. For this purpose, we chose two evaluation datasets that reached a good average rank in \tabref{tab:cross_dataset_gen}. \tabref{tab:svnet_perf_over_iter} reports the results and shows a steady increase for both iterations as our dataset grows. More experiments on the iterative procedure are located in the supplemental material. 

\begin{table}
\begin{center}
\begin{tabular}{|c|c c c c c|}
\hline
Dataset & $\mathcal{D}_0$ & $\mathcal{D}_1$ & $\mathcal{D}_2$ & $\mathcal{D}_3$ & $\mathcal{D}_4$ \\
\hline
\#samples & $302$ & $993$ & $1449$ & $2609$ & $4565$ \\
\textit{RHD} & $0.244$ & $0.453$ & $0.493$  & $0.511$ & $0.518$\\
\textit{PAN} & $0.347$ & $0.521$ & $0.521$  & $0.539$ & $0.562$\\
\hline
\end{tabular}
\caption{Bootstrapping convergence is evaluated by reporting cross-dataset generalization to \textit{RHD} and \textit{PAN}. The measure of performance is AUC, which shows monotonous improvement throughout. }\label{tab:svnet_perf_over_iter}
\vspace{-2em}
\end{center}
\end{table}

\section{Conclusion}

We presented \ourdb, the largest RGB dataset with hand pose and shape labels of real images available to date. We capture this dataset using a novel iterative procedure. The dataset allows us improve generalization performance for the task of 3D hand pose estimation from a single image, as well as supervised learning of monocular hand shape estimation.

To facilitate research on hand shape estimation, we plan to extend our dataset even further to provide the community with a challenging benchmark that takes a big step towards evaluation under realistic \emph{in-the-wild} conditions.

\section*{Acknowledgements}
We gratefully acknowledge funding by the Baden-W\"urttemberg Stiftung as part of the RatTrack project. Work was partially done during Christian’s internship at Adobe Research.

\section{Cross-dataset generalization}
In this section we provide additional information on the single view pose estimation network used in the experiment and other technical details.
The datasets used in the experiment show slight differences regarding the hand model definition: Some provide a keypoint situated at the wrist while others define a keypoint located on the palm instead. To allow a fair comparison, we exclude these keypoints, which leaves $20$ keypoints remaining for the evaluation.
In the subsequent sections we, first provide implementation details in \ref{sec:impl_details}, which includes hyperparameters used and the network architecture. Second, we analyze the influence that using a pretrained network has on the outcome of the experiment in \ref{sec:pretrained_cross_dataset}.

\subsection{Implementation details}\label{sec:impl_details}
We chose our hyper parameters and architecture similar to \cite{iqbal2018hand}. The network consists of an encoder decoder structure with skip connections. For brevity we define the building blocks \textit{Block0} (see \tabref{tab:block0}) \textit{Block1} (see \tabref{tab:block1}), \textit{Block2} (see \tabref{tab:block2}), \textit{Block3} (see \tabref{tab:block3}) and \textit{Block4} (see \tabref{tab:block4}). Using these, the network is assembled according to \tabref{tab:our_arch}. All blocks have the same number of channels for all convolutions throughout. An exception is \textit{Block4}, which has $128$ output channels for the first two and $42$ for the last convolution. The number $42$ arises from $21$ keypoints we estimate 2D locations and depth for. Skip connections from \textit{Block1} to \textit{Block3} always branch off after the last convolution (id $3$) of \textit{Block1} using the respective block that has the same spatial resolution. 

We train the network for \SI{300}{k} iterations with a batch size of $16$. For optimization we use the Momentum solver with an initial learning rate of $0.001$ and momentum of $0.9$. Learning rate is lowered to $0.0001$ after iteration \SI{150}{k}.

\begin{table}[htb!]
\begin{center}
\begin{tabular}{|c|c|c|}
\hline
id & Name & Dimensionality\\
\hline\hline
 & Input image & $128 \times 128 \times 3$\\
1 & Block0 & $128 \times 128 \times 64$\\
2 & Block1 & $64 \times 64 \times 128$\\
3 & Block1 & $32 \times 32 \times 128$\\
4 & Block1 & $16 \times 16 \times 128$\\
5 & Block1 & $8 \times 8 \times 128$\\
6 & Block1 & $4 \times 4 \times 128$\\
7 & Block1 & $2 \times 2 \times 128$\\
8 & Block2 & $4 \times 4 \times 256$\\
9 & Block3 & $8 \times 8 \times 128$\\
10 & Block3 & $16 \times 16 \times 128$\\
11 & Block3 & $32 \times 32 \times 128$\\
12 & Block3 & $64 \times 64 \times 128$\\
13 & Block3 & $128 \times 128 \times 128$\\
14 & Block4 & $128 \times 128 \times 42$\\
\hline
\end{tabular}
\caption{Our single view network architecture used for 3D pose estimation in the cross-dataset generalization experiment.
}\label{tab:our_arch}
\end{center}
\end{table}

\begin{table}[htb!]
\begin{center}
\begin{tabular}{|c|c|c|c|}
\hline
id & Name & Kernel & Stride\\
\hline\hline
1 & Conv. + ReLU & $3 \times 3$ & $1$\\
2 & Avg. Pool & $4 \times 4$ & $1$\\
3 & Conv. + ReLU & $3 \times 3$ & $1$\\
4 & Avg. Pool & $4 \times 4$ & $1$\\
\hline
\end{tabular}
\caption{Block0.
}\label{tab:block0}
\end{center}
\end{table}

\begin{table}[htb!]
\begin{center}
\begin{tabular}{|c|c|c|c|}
\hline
id & Name & Kernel & Stride\\
\hline\hline
1 & Conv. + ReLU & $3 \times 3$ & $1$\\
2 & Avg. Pool & $4 \times 4$ & $2$\\
3 & Conv. + ReLU & $3 \times 3$ & $1$\\
\hline
\end{tabular}
\caption{Block1.
}\label{tab:block1}
\end{center}
\end{table}

\begin{table}[htb!]
\begin{center}
\begin{tabular}{|c|c|c|c|}
\hline
id & Name & Kernel & Stride\\
\hline\hline
1 & Conv. + ReLU & $3 \times 3$ & $1$\\
2 & Upconv. & $4 \times 4$ & $2$\\
\hline
\end{tabular}
\caption{Block2.
}\label{tab:block2}
\end{center}
\end{table}

\begin{table}[htb!]
\begin{center}
\begin{tabular}{|c|c|c|c|}
\hline
id & Name & Kernel & Stride\\
\hline\hline
1 & Concat(Input, Block1 skip ) & - & -\\
2 & Conv. + ReLU & $1 \times 1$ & $1$\\
3 & Conv. + ReLU & $3 \times 3$ & $1$\\
4 & Upconv. & $4 \times 4$ & $2$\\
\hline
\end{tabular}
\caption{Block3.
}\label{tab:block3}
\end{center}
\end{table}

\begin{table}[htb!]
\begin{center}
\begin{tabular}{|c|c|c|c|}
\hline
id & Name & Kernel & Stride\\
\hline\hline
1 & Conv. + ReLU & $7 \times 7$ & $1$\\
2 & Conv. + ReLU & $7 \times 7$ & $1$\\
3 & Conv. & $7 \times 7$ & $1$\\
\hline
\end{tabular}
\caption{Block4.
}\label{tab:block4}
\end{center}
\end{table}

\subsection{Pretrained network}\label{sec:pretrained_cross_dataset}
We provide an additional version of the proposed cross-dataset generalization experiment, which shows the influence of using a pretrained network. For this purpose, we use a \textit{ImageNet} pretrained \textit{ResNet50} backbone, which we train to learn a direct mapping from images to normalized 3D pose. From the original \textit{ResNet50} we use the average pooled final features and process them using $2$ fully connected layers with $2048$ neurons each using ReLU activations and a final linear fully connected layer outputting the $63$ parameters (= $21 \times$ 3D coordinates). Hyperparameters are identical to \ref{sec:impl_details}, except for the use of \textit{ADAM} solver and an initial learning rate of $10^{-5}$, which is lowered to $10^{-6}$ after iteration \SI{150}{k}.
The results are presented in \tabref{tab:cross_dataset_gen_pretrained}, which shows that the average ranks are mostly unchanged compared to the results reported in the main paper. We witness a tendency of lower performance on the respective evaluation set, but better generalization to other datasets.

\begin{table*}[htb!]
\begin{center}
\begin{tabular}{|c|c|c|c|c|c|c|c|c||c|}
\hline
\diagbox[width=5em]{train}{eval} & STB & RHD & GAN & PAN & LSMV & FPA & HO3D & \textbf{Ours} & \thead{Average\\Rank}\\
\hline\hline
STB &\fr{0.687}& $0.247$  & $0.151$  & $0.263$  & $0.220$  & $0.138$  & $0.207$  & $0.244$ & $5.6$\\
RHD &\tr{0.480}&\fr{0.697}& 0.200 & \tr{0.353}& $0.490$ & $0.156$  & \fr{0.417}  &\se{$0.403$}& $2.9$\\
GAN & $0.184$  & $0.198$  &\fr{0.624}& $0.217$  & $0.229$  & $0.182$  & $0.188$  & $0.233$ & $5.8$\\
PAN & $0.447$  &\tr{0.367}& \se{0.221}  &\fr{0.632}& \tr{0.454}  &\tr{0.205}& $0.264$  & \tr{0.345} & $3.0$\\
LSMV & $0.242$  & $0.286$  & $0.199$  & $0.226$  &\fr{0.640}& $0.162$  & \tr{0.283}  & $0.307$ & $4.4$\\
FPA & $0.186$  & $0.178$  & $0.156$  & $0.206$  & $0.197$  &\fr{0.705}& $0.162$  & $0.239$ & $6.6$\\
HO3D & $0.254$  & $0.311$  & $0.198$  & $0.206$  & $0.338$  & $0.148$ & -  & $0.313$ & $5.4$\\
\textbf{Ours}&\se{0.520}&\se{0.399}&\tr{0.205}&\se{0.395}&\se{0.509}&\se{0.208}& \se{0.416} &\fr{0.523} & $2.0$\\
\hline
\end{tabular}
\vspace{-.2em}
\caption{
This table shows, cross-dataset generalization measured as area under the curve (AUC) of percentage of correct keypoints following \cite{zb2017hand}. In contrast to the table reported in the main paper an ImageNet pretrained \textit{ResNet50} network is used for direct regression of normalized 3D pose. Entries are marked if they rank: \textbf{first}, \textcolor{blue}{second} or \textcolor{cyan}{third} for each dataset.
}\label{tab:cross_dataset_gen_pretrained}
\end{center}
\vspace{-2em}
\end{table*}

\section{MVNet}
\subsection{Training loss}
We experimented with different losses for training \textit{MVNet} and witnessed large differences regarding their applicability to our problem. In addition to the loss described in the main paper, which we refer to as \textit{Scorevolume} loss, we used the \textit{Softargmax} loss formulation~\cite{iqbal2018hand}. As reported in literature, we find that \textit{Softargmax} achieves better results for keypoint estimation, but that the respective score for each prediction is less meaningful as a predictor of the expected error.

\begin{figure}
\centering
\small
    \includegraphics[width=1.0\columnwidth, height=.7\columnwidth]{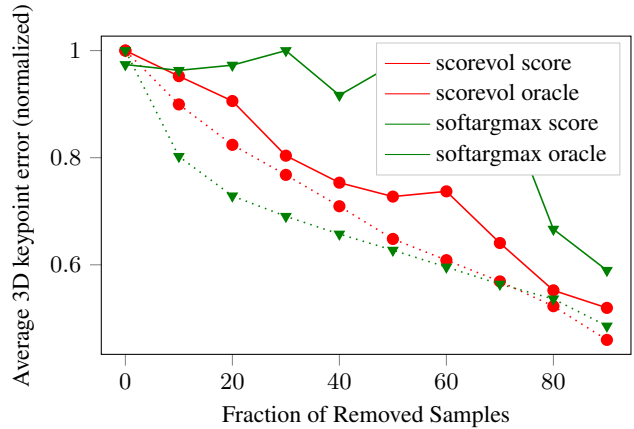}
\caption{Shown is the average predictions 3D error of \textit{MVNet} on a given dataset over the sparsification rate. When moving along the x-axis from left to right the remaining evaluation set gets smaller and the y-axis reports the error upon the remaining dataset. For each approach the resulting curves are shown, when the prediction score is used as solid lines, and when the ground truth error is used as dashed lines, which we refer to as oracle. A good score yields a line that stays close to the oracle line, which shows that \textit{Scorevolume} trained networks learn much more meaningful scores than \textit{Softargmax}.}\label{fig:sparse_error}
\end{figure}

In both cases we define the score $c$ of an prediction that \textit{MVNet} makes as reported in the main paper and use the latent heat map of the \textit{Softargmax} loss to calculate $c$. To analyze the relation between prediction score $c$ and the expected error of the final prediction, we follow the methodlogy described in detail in \cite{ilg2018uncertainty}, and report sparsification curves in \figref{fig:sparse_error}. This plot analyses how the prediction error on a given dataset evolves by gradually removing uncertain predictions, as measured by the prediction score $c$. If the prediction score is a good proxy for the prediction error the curves should monotonically decrease to zero, because predictions with low score should identify samples with high error. The oracle curve shows the ideal curve for a respective loss and is created by accessing the ground truth error instead of using the prediction score, \ie one is always removing the predictions with the largest error.

\figref{fig:sparse_error} shows that score of the \textit{Scorevolume} loss shows much better behavior, because it stays fairly close to its oracle line. Which is in contrast to the \textit{Softargmax} loss. We deduct from this experiment, that the scores that arise when training on a \textit{Scorevolume} represent a more meaningful measure of the algorithms uncertainty and therefore it should be used for our labeling procedure.

\subsection{Implementation details}
The part of network for 2D feature extraction is initialized with the network presented by Simon \etalcite{simon2017hand}. We use input images of size $224 \times 224$, that show hand cropped images. For hand cropping we use a MobileNet architecture \cite{howard2017mobilenets} that is trained on Egohands \cite{bambach_lending_2015} and finetuned on a small, manually labeled, subset of our data. From the 2D CNN we extract the feature encoding $\bm{f}_i$ of dimension $28 \times 28 \times 128$ after $12$ convolutional layers, which is unprojected into a $64 \times 64 \times 64 \times 128$ voxel grid $\bm{F}_i$ of size $0.4$ meters. The voxel grid is centered at a 3D point hypothesis that is calculated from triangulating the detected hand bounding box centers we previously used for image cropping.

For joining the unprojected information from all cameras we average $F_i$ over all views $i$ and use a U-Net \cite{ronneberger2015u} like encoder-decoder architecture in 3D. We train the network for \SI{100}{k} training steps using \textit{ADAM} solver. Batch size is $8$ for the 2D CNNs and $1$ for the 3D CNN. \textit{Scorevolume} loss is used with ground truth Gaussian targets having standard deviation of $2$ voxel units.

\section{Extended Evaluation of iterative procedure}

\begin{figure}
\centering
    \includegraphics[width=.9\columnwidth, height=0.7\columnwidth]{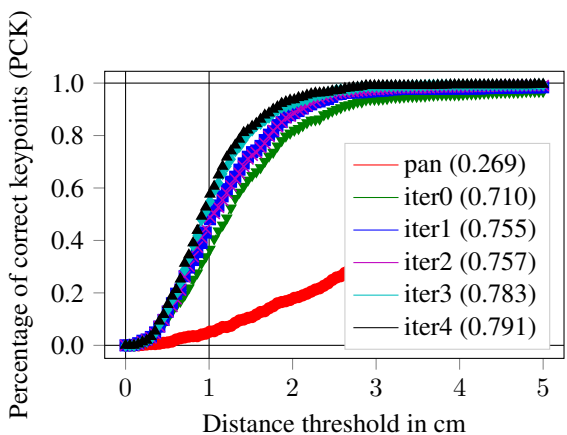}
\caption{We show performance measured as 3D PCK of \textit{
MVNet} over different iterations of our procedure. It shows that training \textit{MVNet} only on the Panoptic dataset is not sufficient to generalize to our data an that each iteration improves performance. In brackets the 3D AUC is reported.}\label{fig:mvnet_over_iter}
\end{figure}

In \figref{fig:mvnet_over_iter}, we show how the 3D keypoint estimation accuracy of \textit{MVNet} evolves over iterations. For comparison, we also shown how \textit{MVNet} performs when trained on the Panoptic (\textit{PAN}) dataset alone. While this gives insufficient performance, joint training on \textit{PAN} and our dataset yields a large gain in performance in the first iteration and every other iteration provides further improvement.

\section{Image Compositing}
\begin{figure*}[!t] 
\centering
\begin{minipage}{1.0\linewidth}
\begin{tabular}{@{}c@{}c@{}@{}c@{}c@{}c@{}}
 \includegraphics[width=.2\linewidth]{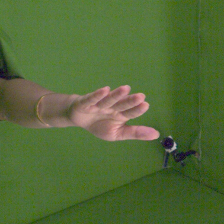}& 
 \includegraphics[width=.2\linewidth]{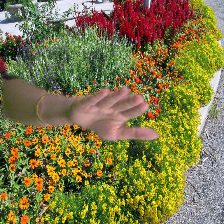}&
 \includegraphics[width=.2\linewidth]{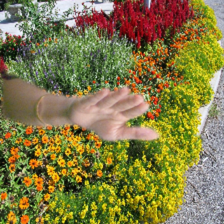}&
 \includegraphics[width=.2\linewidth]{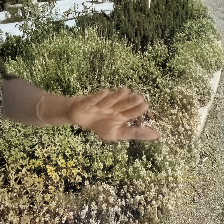}&
 \includegraphics[width=.2\linewidth]{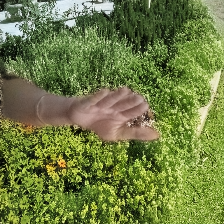}\\ [-\dp\strutbox]
 \end{tabular}
\end{minipage}
\caption{Visualizations of our explored post processing options on the same sample. From left to right: Original frame, \textit{Cut\&Paste}, Harmonization \cite{tsai2017deep}, Colorization Auto \cite{zhang2017real}, Colorization: Sample \cite{zhang2017real}.
}
\label{fig:postproc_variants}
\end{figure*}

Here we study different methods for post processing our recorded images in order to improve generalization of composite images. Green screen indicates that images are used as is \ie no additional processing step was used. \textit{Cut\&Paste} refers to blending the original image with a randomly sampled new background image using the foreground segmentation mask as blending alpha channel. Harmonization is a deep network based approach presented by Tsai \etalcite{tsai2017deep}, which should improve network performance on composite images. Additionally, we experimented with the deep image colorization approach by Zhang \etalcite{zhang2017real}. For this processing step we convert the composite image of the \textit{Cut\&Paste} method into a grayscale image and input it in the colorization method. Here we have the options \textit{Auto}, in which the network hallucinates all colors and \textit{Sample} where we provide the network with the actual colors in $20$ randomly chosen sample points on each foreground and background. Examples of images these approaches yield are shown in \figref{fig:postproc_variants}. \tabref{tab:post_proc_iqbal} reports results for the network described in \ref{sec:impl_details} and \tabref{tab:post_proc_baseline} shows results when the pretrained baseline is used instead. The two tables show, that post processing methods are more important when networks are trained from scratch. In this case, \tabref{tab:post_proc_iqbal} shows that using each of the processing options yields roughly the same gain in performance and using all options jointly performs best. This option is chosen for the respective experiments in the main paper. When a pretrained network is used \tabref{tab:post_proc_baseline} reports already good results, when the network is only trained on green screen images. Interestingly, in this scenario the more elaborate post processing methods yield only a minor gain compared to the \textit{Cut\&Paste} strategy. We hypothesize these results are related to a significant level of robustness the pretrained weights possess.
Please note that we can't use these algorithms in a similar manner for datasets that don't provide segmentation masks of the foreground object. Only \textit{RHD} provides segmentation masks, which is why we show the influence the discussed processing methods have on its generalization. \tabref{tab:post_proc_rhd} shows that the discussed methods don't improve performance for \textit{RHD} trained networks the same way. The results indicate that these strategies should not been seen as general data augmentation, but rather specific processing steps to alleviate the problem of green screen color bleeding we witness on our training dataset.

\begin{table}[htb!]
\begin{center}
\begin{tabular}{|l|c|c|}
\hline
\diagbox[width=12em]{method}{eval} & RHD & \textbf{Ours}\\
\hline\hline
Green Screen & $0.246$ & $0.440$ \\
Cut\&Paste & $0.350$ & $0.508$ \\
Harmonization \cite{tsai2017deep} & $0.443$ & $0.628$ \\
Colorization Auto \cite{zhang2017real} & $0.458$ & $0.634$ \\
Colorization Sample \cite{zhang2017real}  & $0.494$ & $0.643$ \\
Joint & $\mathbf{0.518}$ & $\mathbf{0.678}$ \\
\hline
\end{tabular}
\caption{Network architecture as used in the main paper. When training networks from scratch, it is important to introduce background variation by inserting random backgrounds into the green screen recordings. Using post processing algorithms improves generalization.
}\label{tab:post_proc_iqbal}
\end{center}
\end{table}

\begin{table}[htb!]
\begin{center}
\begin{tabular}{|l|c|c|}
\hline
\diagbox[width=12em]{method}{eval} & RHD & \textbf{Ours}\\
\hline\hline
Green Screen & $0.338$ & $0.436$ \\
Cut\&Paste & $0.386$ & $0.468$ \\
Harmonization \cite{tsai2017deep} & $\mathbf{0.416}$ & $0.513$ \\
Colorization Auto \cite{zhang2017real} & $0.368$ & $0.478$ \\
Colorization Sample \cite{zhang2017real} & $0.381$ & $0.496$ \\
Joint & $0.399$ & $\mathbf{0.523}$ \\
\hline
\end{tabular}
\caption{Instead of the network architecture used in the main paper, we use a \textit{ResNet50} Baseline as described in \ref{sec:pretrained_cross_dataset}. When the network is initialized with weights that already show a certain level of robustness the importance of background removal and recombination with post processing is less pronounced, but still improves results substantially.
}\label{tab:post_proc_baseline}
\end{center}
\end{table}

\begin{table}[htb!]
\begin{center}
\begin{tabular}{|l|c|c|}
\hline
\diagbox[width=12em]{method}{eval} & RHD & \textbf{Ours}\\
\hline\hline
Original & $\mathbf{0.767}$ & $0.508$ \\
Harmonization \cite{tsai2017deep} & $0.723$ & $\mathbf{0.517}$ \\
Colorization Auto \cite{zhang2017real} & $0.726$ & $0.472$ \\
Colorization Sample \cite{zhang2017real} & $0.748$ & $0.501$ \\
Joint & $0.756$ & $0.514$ \\
\hline
\end{tabular}
\caption{Network architecture as used in the main paper, but instead of training on our dataset we train on \textit{RHD} and apply the same post processing methods to it. We chose \textit{RHD} as reference because it is the only dataset that also provides foreground segmentation masks, which are needed for the processing methods. The table shows that none of them yields clear improvements over using the original unaltered \textit{RHD} dataset for training. 
}\label{tab:post_proc_rhd}
\end{center}
\end{table}

\section{\ourdb~details}
For the dataset we recorded $32$ people and asked them to perform actions in front of the cameras. The set of non object actions included: signs from the american sign language, counting, move their fingers to their kinematic limits. The set of objects contains different types of workshop tools like drills, wrenches, screwdrivers or hammers. Different types of kitchen supply were involved as well, f.e. chopsticks, cutlery, bottles or BBQ tongs. These objects were either placed into the subjects hand from the beginning of the recording or hang into our setup and we recorded the process of grabbing the object.

Actions that contain interaction with objects include the following items: Hammer, screwdrive, drill, scissors, tweezers, desoldering pump, stapler, wrench, chopsticks, caliper, power plug, pen, spoon, fork, knive, remote control, cream tube, coffee cup, spray can, glue pistol, frisbee, leather cover, cardboard box, multi tool and different types of spheres (\fe apples, oranges, styrofoam). The action were selected such that all major grasp types were covered including power and precision grasps or spheres, cylinders, cubes and disks as well as more specialized object specific grasps.

\begin{figure*}[!t] 
\centering
\begin{minipage}{1.0\linewidth}
\begin{tabular}{@{}c@{}c@{}c@{}c@{}}
 \includegraphics[width=.25\linewidth]{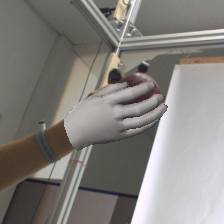}& 
 \includegraphics[width=.25\linewidth]{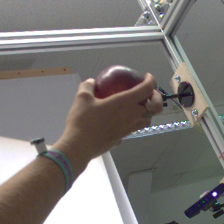}&
 \includegraphics[width=.25\linewidth]{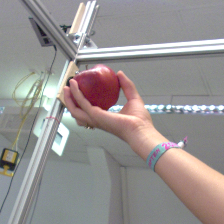}&
 \includegraphics[width=.25\linewidth]{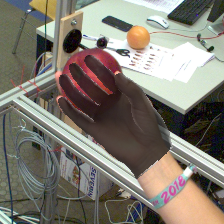}\\ [-\dp\strutbox]
 
 \includegraphics[width=.25\linewidth]{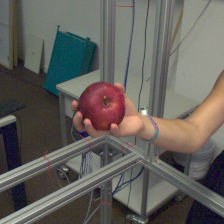}&
 \includegraphics[width=.25\linewidth]{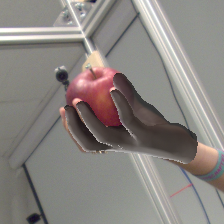}&
 \includegraphics[width=.25\linewidth]{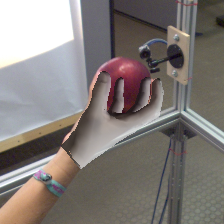}&
 \includegraphics[width=.25\linewidth]{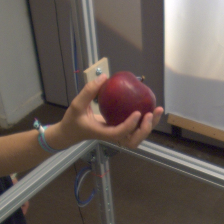}\\
\end{tabular}
\end{minipage}
\caption{Here we show \textbf{one sample} of our dataset, which consists of $\mathbf{8}$ \textbf{images} that are recorded at the same time instance. We overlay the shape label found with our method in some of the images.
}
\label{fig:dataset_single_sample_examples}
\end{figure*}

\begin{figure*}[!t] 
\centering
\begin{minipage}{1.0\linewidth}
\begin{tabular}{@{}c@{}c@{}c@{}@{}c@{}c@{}c@{}}
 \includegraphics[width=.16\linewidth]{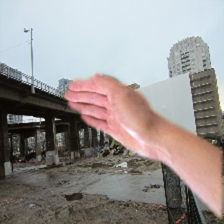}& 
 \includegraphics[width=.16\linewidth]{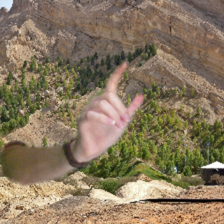}&
 \includegraphics[width=.16\linewidth]{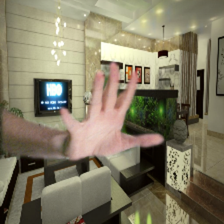}&
 
 \hspace{.014\linewidth}
 
 \includegraphics[width=.16\linewidth]{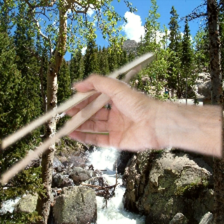}&
 \includegraphics[width=.16\linewidth]{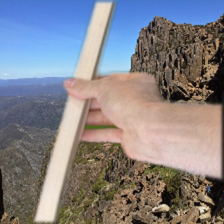}&
 \includegraphics[width=.16\linewidth]{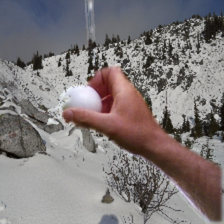}\\ [-\dp\strutbox]
 
 \includegraphics[width=.16\linewidth]{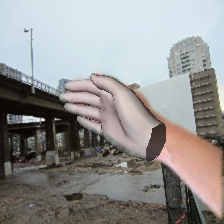}&
 \includegraphics[width=.16\linewidth]{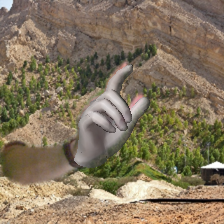}&
 \includegraphics[width=.16\linewidth]{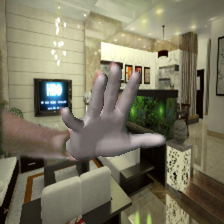}&
 
 \hspace{.014\linewidth}
 
 \includegraphics[width=.16\linewidth]{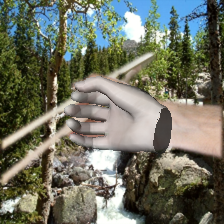}&
 \includegraphics[width=.16\linewidth]{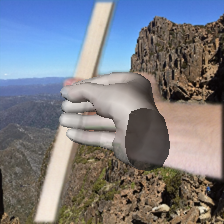}&
 \includegraphics[width=.16\linewidth]{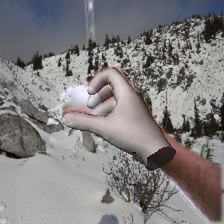}\\
\end{tabular}
\end{minipage}
\caption{Examples from our proposed dataset showing images and hand shape annotations. It shows the final images that are used for training of single view methods with the original background replaced and \cite{tsai2017deep} applied for post processing.
}
\label{fig:more_dataset_examples}
\end{figure*}

These recordings form the basis we run our iterative labeling procedure on, that created the dataset presented in the main paper. Some examples of it are shown in \figref{fig:more_dataset_examples}. \figref{fig:dataset_single_sample_examples} shows one dataset sample containing $8$ images recorded at a unique time step from the $8$ different cameras involved in our capture setup. One can see that the cameras capture a broad spectrum of viewpoints around the hand and how for different cameras different fingers are occluded. Our datasets shape annotation is overlayed in half of the views.

\begin{figure*}[!t] 
\centering
\begin{minipage}{1.0\linewidth}
\begin{tabular}{@{}c@{}c@{}c@{}@{}c@{}c@{}c@{}}
 \includegraphics[width=.16\linewidth]{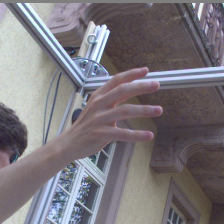}& 
 \includegraphics[width=.16\linewidth]{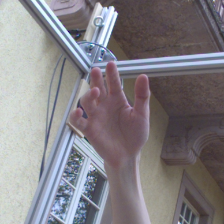}&
 \includegraphics[width=.16\linewidth]{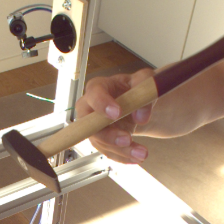}&
 
 \hspace{.014\linewidth}
 
 \includegraphics[width=.16\linewidth]{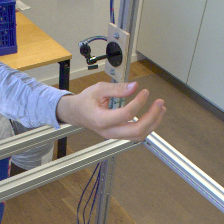}&
 \includegraphics[width=.16\linewidth]{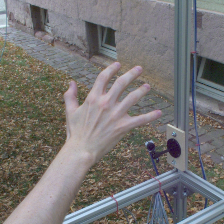}&
 \includegraphics[width=.16\linewidth]{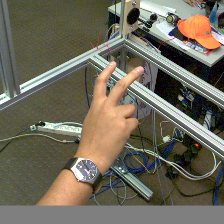}\\ [-\dp\strutbox]
 
 \includegraphics[width=.16\linewidth]{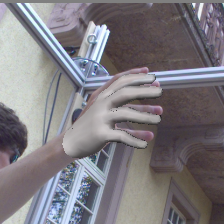}&
 \includegraphics[width=.16\linewidth]{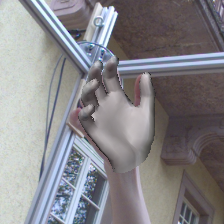}&
 \includegraphics[width=.16\linewidth]{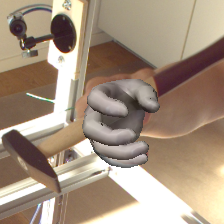}&
 
 \hspace{.014\linewidth}
 
 \includegraphics[width=.16\linewidth]{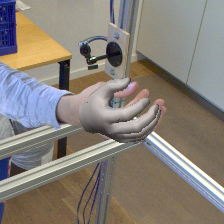}&
 \includegraphics[width=.16\linewidth]{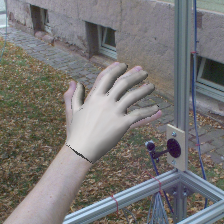}&
 \includegraphics[width=.16\linewidth]{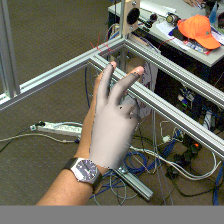}\\
\end{tabular}
\end{minipage}
\caption{More qualitative examples of predicted hand shapes that our single view network makes. Again, we don’t apply any alignment of the predictions with respect to the ground truth, which explains minor miss alignment whereas the hand articulation is captured correctly.
}
\label{fig:qualitative_examples_svnet}
\end{figure*}

Furthermore, we provide more qualitative examples of our single view shape estimating network in \figref{fig:qualitative_examples_svnet}.

Distributions across genders and ethnicity's are reported in \tabref{tab:gender_eth_distribution}, whereas \tabref{tab:sample_distribution} shows the distribution of labeled samples across gender and object interaction. 

\begin{table}
\begin{center}
\resizebox{\columnwidth}{!}{%
\begin{tabular}{|l|r|r|r|r|r|}
\hline
Part & \thead{Samples \\with object} & \thead{Samples \\w/o object} & male & female & total\\
\hline\hline
 Training (green screen) & $2580$ & $1490$ & $2462$ & $1608$ & $4070$ \\
 Evaluation (total) & $339$ & $156$ & $290$ & $205$ & $495$\\
 Evaluation (plane space office) & $119$ & $40$ & $63$ & $96$ & $159$ \\
 Evaluation (outdoor) & $114$ & $47$ & $88$ & $73$ & $161$ \\
 Evaluation (meeting room) & $106$ & $69$ & $139$ & $36$ & $175$ \\
\hline
\end{tabular}
}%
\caption{\small Distribution of labeled samples in our dataset across different aspects. }\label{tab:sample_distribution}
\end{center}
\end{table}

\begin{table}
\begin{center}
\resizebox{\columnwidth}{!}{%
\begin{tabular}{|l|r|r|}
\hline
Ethnicity & Training & Evaluation\\
\hline\hline
 Caucasian (NA, Europe, ...) & $14$ & $11$ \\
 South Asian (India, Pakistan, ...) & $3$ & $2$ \\
 East Asia (China, Vietnam, ...) & $2$ & $0$ \\
 \hline
 Male subjects & $15$ & $6$ \\
 Female subjects & $9$ & $5$ \\
\hline
\end{tabular}
}%
\caption{\small Gender and ethnicity distribution of recorded subjects in the dataset.}\label{tab:gender_eth_distribution}
\end{center}
\end{table}

The dataset was recorded using the following hardware: Two Basler acA800-510uc and six Basler acA1300-200uc color cameras that were hardware triggered using the GPIO module by numato. The cameras were equipped with fixed lenses with either 4mm or 6mm focal length. The recording setup is approximately forming a cube of edge length $1$m with one of the color cameras being located in each of the corners. The subjects then reached inside the cubicle through one of the cubes' faces, which approximately put their hand at an equal distance to all the cameras. When recording for the evaluation split, we used ambient lighting. To improve lighting during green screen recording there were $4$ powerful LED lights as used during recording with a video camcorder. These allowed to vary in terms of lighting power and light temperature during the recordings.

{\small
\bibliographystyle{ieee_fullname}
\bibliography{bibfile}
}

\end{document}